# Talk Less, Verify More:
# Improving LLM Assistants with Semantic Checks and Execution Feedback

*Regular Paper*

Yan Sun, Ming Cai and Stanley Kok
yan.sun@u.nus.edu caim@u.nus.edu stanley.kok@nus.edu.sg
Department of Information Systems and Analytics
National University of Singapore

## Introduction

Large language model (LLM)-based assistants are increasingly integrated into enterprise systems to support data access, automate structured workflows, and enhance productivity across a range of business functions (Chau and Xu 2025). These tools are rapidly being adopted in domains such as customer relationship management (CRM), marketing, finance, healthcare, law, and education. In CRM, for example, LLMs handle customer inquiries and generate personalized support using organizational knowledge via frameworks like LangChain and LangGraph (Lewis et al. 2020). In marketing, LLMs produce tailored advertising content, generate campaign insights, and predict customer preferences through analysis of structured and unstructured data (Kshetri et al. 2024). More broadly, enterprises are beginning to build multi-agent workflows that leverage LLMs for planning, reasoning, and tool-based interaction, enabling automation of increasingly complex decision-making tasks (Akkil et al. 2024).Across these domains, business users are shifting toward natural language interfaces for analytics-related tasks. This shift marks the rise of conversational business analytics (CBA), a new paradigm in enterprise information systems that enables users to interact with structured data through conversational interfaces (Alparslan 2025). Systems such as Amazon Q Business and Microsoft Copilot exemplify this shift, allowing users to filter datasets, compute KPIs, and analyze trends using natural language (Bruhin 2024). These tools aim to broaden access and improve efficiency, particularly for non-technical decision-makers (Vroegindeweij et al. 2024).

Despite these advances, current CBA systems exhibit a critical shortcoming: they lack robust mechanisms for verifying that generated outputs are semantically aligned with user intent and logically executable. As shown in Figure 1, most systems operate in a linear fashion, which converts a user's query into code (e.g., SQL/Python) via the LLM assistant, executes that code to generate results, and returns those results to the user. However, this pipeline assumes the correctness of both the generated code and the resulting output. It places the burden of validation squarely on the user, who may not possess the technical expertise to identify subtle but significant errors. Misinterpretations in query logic, such as incorrect filters or aggregations, can easily lead to flawed analyses or misleading dashboards (Chen and Chan 2024). These hidden errors introduce verification risk, where undetected faults propagate through business processes, undermining trust in AI-assisted decision-making (Eigner and Händler 2024).

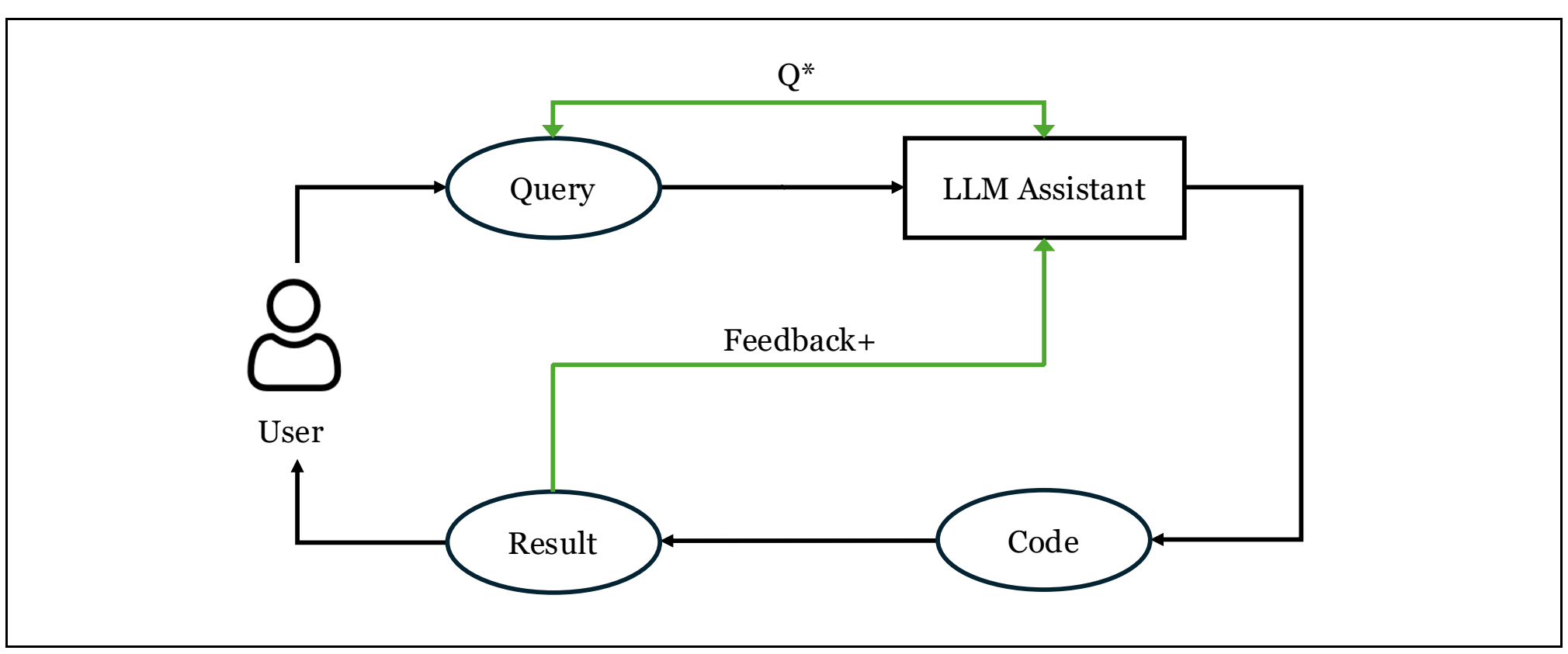

**Figure 1. Overview of Proposed Verification Mechanisms (in Green)**

In practice, fragile validation leads to repeated user clarifications and cross-checks—the "talk-more" effect—which shifts quality control to the user and slows decision cycles. The design gap is clear: current assistants generate fluent code, but verification is weak. In enterprise systems, even small SQL or logic errors can distort KPIs or breach compliance. Unlike open-domain tasks, enterprise analytics require fully verified and reproducible outputs. Thus, verification must be built into the assistant itself, not left to users. To address this gap, we propose a new design principle: *Talk Less, Verify More*. Rather than relying on the user for validation, we shift verification responsibilities into the system by embedding two mechanisms, termed Q* and Feedback+, into the LLM interaction loop. Q* performs semantic verification by reverse translating generated code back into natural language to assess alignment with the original query. Feedback+ introduces execution-grounded validation, assessing code quality based on its runtime behavior and logical coherence. As shown in green in Figure 1, these verification mechanisms enable the assistant to refine its outputs before presenting them to the user.

We evaluate this approach across three benchmark datasets: Spider and Bird for text-to-SQL generation, and GSM8K for mathematical reasoning and Python generation. These datasets simulate core enterprise use cases that demand both semantic precision and logical correctness. Empirical results show that integrating Q* and Feedback+ leads to substantial reductions in both error rates and task completion times, particularly in tasks where user intent can be expressed in semantically structured forms. Further analysis reveals that while semantic matching is reliably handled by modern LLMs, the fidelity of reverse translation remains a limiting factor for broader generalization—especially in domains with implicit or abstract logic.

## Related work

### *Enterprise GenAI Assistant*

The use of Generative AI (GenAI) assistants in enterprises is growing, driven by their ability to automate knowledge work and improve efficiency (Chau and Xu 2025). Large-scale evidence reports sizable productivity gains: a field experiment with over 5,000 customer support agents found a 15% average increase, with the largest gains for less experienced workers (Brynjolfsson et al. 2025). Trane Technologies reported improvements of 3.3%–69% across tasks such as email drafting and data summarization (Freeman et al. 2024). Goldman Sachs deployed "GS AI Assistant" to thousands of employees for faster document creation, code generation for financial models, and report translation, which streamlined core processes and supported ideation (Business Insider 2025). Despite these results, current systems struggle with consistent, context-specific accuracy and low latency, especially in multi-turn settings where retrieval and ranking remain difficult. Many organizations also rely on custom solutions and manual prompt engineering, which can erode time savings. Our study addresses these gaps by proposing a design and evaluation framework for enterprise GenAI assistants that prioritizes response accuracy and time efficiency in high-value internal tasks, offering practical, reproducible guidance grounded in prior empirical and deployment evidence.

### *LLM Planning*

Recent work shows that large language models (LLMs) can handle multi-step planning using methods such as reranking (Li et al. 2023), iterative correction (Madaan et al. 2024), and tree search (Yao et al. 2024), typically within an Actor-Critic setup. Note that there are other framework with additional module, for example, Reflexion introduces a separate Self-Reflection model to generate verbal feedback, but beyond the scope of this study. In the Actor-Critic setup, the Actor produces multiple candidate outputs (e.g., code snippets or stepwise explanations), and the Critic—another LLM or a task-specific module—scores them. In iterative correction, the Actor creates initial plans that the Critic ranks; the top plan is fed back to the Actor for revision until a score threshold is met or scores stop improving for three rounds. However, interactive refinement often falls short (Gu et al. 2023; West et al. 2023), partly because the Critic may misjudge plan alignment with the ground truth. Prior studies (Huang et al. 2023; Wang et al. 2022) show that valid plans can be rejected early, leading to cascading errors, a problem that is acute in enterprise use where errors undermine trust (Saffarizadeh et al. 2024). Although recent work improves planning via in-loop ground-truth checks (Chen et al., 2024; Huang et al., 2023), it remains confined to self-correction. We enhance LLM planning by adding semantic and execution-based verification beyond this loop.

## Our Method

To ensure reliability, enterprise LLM assistants often require users to validate whether the generated code (e.g., SQL) accurately reflects their original intent. However, manual validation is prone to error, particularly when users lack the technical expertise to interpret complex logic (Alparslan 2025). We propose two verification mechanisms for improving LLM assistants: (i) semantic consistency checking between generated code and the input query, and (ii) leveraging execution feedback to refine the Actor. These mechanisms are complementary to each other.

### ***Q*: Reverse Translation and Semantic Matching***

Our first mechanism is an automated, model-internal mechanism for validating semantic alignment between the input query and the generated code. Specifically, it consists of two stages: reverse translation and semantic matching.

In the first stage, reverse translation, the generated code is translated back into a natural language query. This step makes the model's interpretation of the task explicit and comparable to the original input. For example, in a text-to-SQL setting, we prompt the Critic LLM as follows[1]:

```
Given the database schema and SQL below, generate the corresponding question
that the SQL tries to solve (ending with a question mark).
-- Database: {}
-- SQL:{}
-- Question:
```

The output of this process is a natural language question Q* that reflects the meaning of the generated code. In the second stage, semantic matching, we compare Q* to the original query Q by prompting the Critic to assess whether they express the same intent:

```
Answer the following Yes/No question: Do these two questions have the same
meaning?
-- Question 1: {}
-- Question 2: {}
-- Answer:
```

We set the discrimination score to be the probability of Yes being generated as the next token. A high score indicates strong semantic alignment; otherwise, the code is less likely to solve the user's query. Thus, in the next round, the top-ranked candidates are then used to guide the Actor in the next round, using a corrective prompt:

```
Given a database schema and a question in natural language, correct the
candidate SQL query based on the execution result and generate a fixed SQL
query.
```

[1] We present only the SQL-related prompt due to space constraints.

```
-- Database: {}
-- Question: {}
-- Candidate SQL: {}
Fixed SQL:
```

***Feedback+: Execution Results for Actor***

When executing the generated code, it is possible that the errors occur with feedback in detail. Instead of the manual check the execution by users, the second mechanism, termed as Feedback+, reduces the workload by directly feeding the execution results into the Actor.

The result of its execution may include whether it succeeded or failed, and any error messages. The prompt guides the LLM to take this feedback and generate a corrected code, making it more efficient in refining the program iteratively. In the text-to-SQL task, we implement this mechanism using the following prompt template, which captures the database schema, the original query, the generated SQL query, the execution result, and the corrected query as below. *Buggy SQL* denotes a candidate SQL that either fails to execute or executes but does not answer the user's question.

```
Given a database schema and a question in natural language, correct the buggy
SQL query based on the execution result and generate a fixed SQL query.
-- Database: {}
-- Question: {}
-- Buggy SQL: {}
-- Execution Result: {}
-- Fixed SQL:
```

## Experiments and Primary Results

To evaluate the core capabilities required for an Enterprise GenAI Assistant, we focus on two task categories that are highly relevant to enterprise settings: (1) translating user intent into data queries and (2) supporting structured, multi-step reasoning. These tasks reflect essential workflows in data-driven business decision-making. We use three benchmark datasets to evaluate our methods in a controlled, replicable environment. Our evaluation emphasizes two key aspects: performance comparison across diverse tasks and efficiency analysis for real-world applicability.

### *Tasks and Datasets*

**Text-to-SQL Parsing** In many enterprise applications, users—such as analysts, managers, or frontline staff—need to query business data without writing complex code. The ability to convert natural language questions into structured SQL queries is therefore a foundational capability for any GenAI assistant operating on enterprise databases. To simulate this functionality, we adopt two benchmark datasets: (i) **Spider** (Yu et al. 2018), a complex, cross-domain benchmark featuring natural language questions and relational databases, mimicking the variability of enterprise data environments; and (ii) **Bird** (Li et al. 2024), a more recent dataset that evaluates schema linking and understanding in nested and multi-table SQL queries, aligning with realistic enterprise reporting needs. For training, we use the full training split from each dataset. For evaluation, considering resource and budget limitations, we randomly select 400 examples, suitable for enterprise contexts, from the development sets of Spider and Bird.

**Mathematical Reasoning** Enterprises frequently require answers that go beyond data retrieval—such as cost-benefit analysis, forecasting, or scenario planning. These tasks involve multiple steps of logical and numerical reasoning. To model this capability, we use **GSM8K** (Cobbe et al. 2021), a benchmark consisting of grade-school level math word problems. This dataset tests the assistant's ability to generate executable reasoning plans in the form of Python code, simulating enterprise logic that underpins business calculations. For our experiments, we sample 500 questions from the GSM8K development set and follow the "program of thoughts" framework (Chen et al. 2022).

### *Experiment Setup*

We adopt the same iterative-correction loop across Baseline, Q*, and Feedback+. Non-executable outputs are filtered before further processing in all methods.

**Baseline (Iterative Correction)**. We use the actor-critic framework from Chen et al. (2024) as our baseline. It performs up to 10 correction rounds, with early stopping if (i) the best plan achieves a discrimination score above 0.99, or (ii) the score does not improve over three consecutive rounds. CodeLlama-7B is used as the Actor in a zero-shot setting, and GPT-3.5-turbo serves as the Critic, providing feedback to guide revisions.

**Q* mechanism.** We follow the same iterative setup (10 rounds, early stopping, and CodeLlama-7B as Actor) but replace the critic-based feedback with two steps using GPT-3.5-turbo: (1) translation of code to natural language, and (2) semantic matching to assess alignment with the input intent.

**Feedback+ mechanism.** We also use the same iterative correction loop and Actor but replace the critic feedback with execution results. The generated code is run in a compiler, and the execution feedback (e.g., errors or outputs) is used to guide subsequent revisions.

## Results

### *Accuracy Comparison Across Tasks*

We evaluate based on the accuracy, which is the proportion of generated programs that execute successfully and produce the correct output according to ground-truth labels from each dataset. As shown in Table 1, our experimental results demonstrate several key points. Firstly, Feedback+ outperformed Q* in most cases, providing greater improvements over both the baseline and Q*.

This suggests that explicit feedback and iterative refinement are crucial for enhancing model performance, especially in complex tasks. Secondly, the degree of improvement varied across datasets, indicating that the effectiveness of our methods is task-dependent. For instance, while Q* was beneficial for the Spider and Bird datasets, it did not perform well on the GSM8K dataset. This likely occurs because Q verifies alignment with the query's overall semantic meaning, which suits SQL tasks, but fails to ensure the step-by-step logical correctness crucial for GSM8K.

| Dataset | Baseline | Q* | Feedback+ |
|---|---|---|---|
| Spider | 44.30 | 46.00 | 46.50 |
| Bird | 10.33 | 12.33 | 15.00 |
| GSM8K | 29.00 | 22.00 | 41.00 |

**Table 1. Comparison of Accuracy (%) Between Our Methods and the Baseline.**

***Disentangle Q****

While Q* demonstrates strong performance on structured tasks such as Spider and Bird, its effectiveness drops significantly on GSM8K. To better understand which components, drive Q*'s success—and where its current limitations lie—we isolate and evaluate the semantic matching module. We constructed a semantic-matching dataset from the Spider dataset, consisting of 12,000 training and 600 test samples. Each sample comprises a pair of natural language questions and asks whether they express the same intent. Many examples involve subtle distinctions (e.g., "ascending" vs. "descending" order) that require precise semantic comprehension. In a zero-shot setting, GPT-3.5-turbo achieves an impressive 93.50% accuracy on the test set, demonstrating strong generalization without task-specific training. In contrast, a fine-tuned all-MiniLM-L6-v2 model (Reimers and Gurevych 2019) reaches only 53.50% accuracy.

| Model | Accuracy (%) |
|---|---|
| GPT-3.5-turbo | 93.50 |
| all-MiniLM-L6-v2 | 53.50 |

**Table 2. Performance on Q* Semantic Matching Dataset.**

These findings suggest that GPT-3.5-turbo is inherently well-suited for semantic matching tasks, even without fine-tuning. Thus, future efforts to enhance Q* should focus less on the semantic module and more on the reverse translation component. In complex domains like GSM8K, translating structured outputs back into natural language queries remains a significant challenge, often requiring reasoning beyond surface-level alignment. Addressing this bottleneck may yield substantial performance gains.

### *Efficiency Analysis*

**Theoretical Complexity.** All methods follow the same actor–critic loop structure. In each round, interacting with the environment takes $O(E)$ time. The baseline method generates sequences of length $n$ with hidden dimension $d$, with hidden dimension $O(n^2d + E)$. Q adds a semantic validation step but retains the same asymptotic complexity. Feedback+ incorporates execution results of length $p_{exec}$ into the prompt, which increases the input length and results in $O((n + p_{exec})^2 d + E)$ time complexity.

| Baseline | Q* | Feedback+ |
|---|---|---|
| $O(n^2d + E)$ | $O(n^2d + E)$ | $O((n + p_{exec})^2 d + E)$. |

**Table 3. Time complexity of each method under the iterative correction framework.**

**Empirical Runtime.**

We characterize the efficiency gains in terms of total wall-clock time completing all the tasks. As shown in Table 4, the baseline required 29.0 hours on the Spider dataset due to repeated low-quality generations and inefficient revisions, whereas Q* achieved faster convergence, reducing runtime to 13.5 hours by better aligning generation with intent. Feedback+, though slower (18.5 hours), still outperformed the baseline by generating higher-quality outputs with fewer revision rounds. These findings suggest that both Q* and Feedback+ improve planning efficiency in practice. While Feedback+ incurs a higher per-iteration cost, it reduces overall runtime.

| Method | Baseline | Q* | Feedback+ |
|---|---|---|---|
| Runtime (hours) | 29.00 | 13.50 | 18.50 |

**Table 4. Runtime Comparison on the Spider Dataset.**

## Conclusions and Future Work

In this study, we focus on LLM-based assistants for structured enterprise tasks that require both semantic precision and executable correctness. We introduce two mechanisms within an Actor–Critic framework: intent alignment through reverse translation and iterative refinement from execution feedback. Across three datasets, these mechanisms improve accuracy and reduce task completion time, addressing reliability challenges in planning and reasoning. In enterprise settings, the proposed methods are expected to lower analytic risk by verifying both intent and execution, thereby improving decision efficiency with less user intervention. Future work should enhance the reverse translation component and evaluate performance on larger enterprise datasets.